\documentclass[preprint,12pt,authoryear]{elsarticle}

\usepackage{amssymb}
\usepackage{amsmath}
\usepackage{hyperref}
\usepackage{xcolor}
\hypersetup{
    colorlinks=true,   
    linkcolor=blue,    
    citecolor=blue,    
    urlcolor=blue      
}
\usepackage{booktabs}
\usepackage{rotating}
\usepackage{amsfonts}
\usepackage[utf8]{inputenc} 
\usepackage[T1]{fontenc}    
\usepackage{multirow}
\usepackage{float}
\usepackage{rotating}
\usepackage{svg}

\journal{Remote Sensing of Environment}

\begin{document}

\begin{frontmatter}

\title{Neural Processes Maintain Calibrated Biomass Estimates Across Spatiotemporal Gaps and Disturbance}

\author[cam]{Robin Young}
\ead{ray25@cam.ac.uk}

\author[cam]{Srinivasan Keshav}
\ead{sk818@cam.ac.uk}

\affiliation[cam]{organization={Department of Computer Science and Technology},
            institution={University of Cambridge},
            city={Cambridge},
            postcode={CB3 0FD}, 
            state={Cambridgeshire},
            country={UK}}

\begin{abstract}
Monitoring deforestation-driven carbon emissions requires both spatially explicit and temporally continuous estimates of aboveground biomass density (AGBD) with calibrated uncertainty. NASA's Global Ecosystem Dynamics Investigation (GEDI) provides reliable LIDAR-derived AGBD, but its orbital sampling causes irregular spatiotemporal coverage, and occasional operational interruptions, including a 13-month hibernation from March 2023 to April 2024, leave extended gaps in the observational record. Prior work has used machine learning approaches to fill GEDI's spatial gaps using satellite-derived features, but temporal interpolation of biomass through unobserved periods, particularly across active disturbance events, remains largely unaddressed. Moreover, standard ensemble methods for biomass mapping have been shown to produce systematically miscalibrated prediction intervals. To address these gaps, we extend the Attentive Neural Process (ANP) framework, previously applied to spatial biomass interpolation, to jointly sparse spatiotemporal settings using geospatial foundation model embeddings. We treat space and time symmetrically, empirically validating a form of space-for-time substitution in which observations from nearby locations at other times inform predictions at held-out periods. We evaluate performance across three ecologically distinct regions: a tropical deforestation frontier in Guaviare, Colombia; dense Amazonian rainforest in Ucayali, Peru; and semi-arid woodland in Queensland, Australia and stratify results by disturbance intensity to test whether calibration holds where temporal stationarity assumptions are weakest. Our results demonstrate that the ANP produces well-calibrated uncertainty estimates across disturbance regimes, supporting its use in Measurement, Reporting, and Verification (MRV) applications that require reliable uncertainty quantification for forest carbon accounting.
\end{abstract}




\end{frontmatter}


\section{Introduction}

Deforestation is an important source of anthropogenic carbon emissions, contributing approximately 10--20\% of global greenhouse gas output annually \citep{pan2011carbon, pendrill2019deforestation}. Monitoring forest loss and quantifying the associated carbon emissions requires spatially explicit, temporally continuous estimates of aboveground biomass density (AGBD) \citep{baccini2012estimated}. International frameworks for climate mitigation, including the UNFCCC REDD+ mechanism, mandate that participating countries report forest carbon stock changes with quantified uncertainty to meet Measurement, Reporting, and Verification (MRV) standards \citep{iso14064-1-2018}. Unreliable uncertainty estimates undermine this process as overconfident bounds can overstate the precision of carbon credits, while underconfident bounds waste resources on unnecessary verification \citep{haya2020managing}.

NASA's Global Ecosystem Dynamics Investigation (GEDI) mission provides LIDAR estimates of AGBD at $\sim$25\,m footprint resolution from the International Space Station \citep{dubayah2022gedi}. However, GEDI's orbit and instrument cycle produce irregular spatiotemporal sampling. Footprints are separated by $\sim$600\,m in criss-crossing orbital patterns, and temporal revisit intervals vary depending on latitude and operational constraints \citep{dubayah2020global}. When disturbance events occur between observation periods, they may be invisible to direct measurement. A tract of forest cleared in early 2021, for example, might be captured by GEDI acquisitions in 2020 and 2022 but not during the year of disturbance itself \citep{holcomb2024repeat}.

Recent work has demonstrated that GEDI's spatial gaps can be filled by training machine learning models on satellite-derived features, producing wall-to-wall biomass maps from sparse footprints \citep{shendryk2022fusing, sialelli2025agbd, nascetti2023biomassters}. However, the temporal dimension of predicting biomass during periods without GEDI coverage has received less attention. Temporal approaches to biomass estimation from optical and SAR time series have been explored using Landsat trajectories and change detection algorithms \citep{arevalo2023continuousAGB}, but these produce relative change indices rather than calibrated absolute AGBD estimates with quantified uncertainty, which is what MRV frameworks require. 

Prior work \citep{young2026interpolation} demonstrated that standard ensemble methods for spatial biomass mapping (Random Forest, XGBoost) produce systematically miscalibrated uncertainty estimates, with 1$\sigma$ prediction intervals capturing as few as 19\% of held-out observations instead of the nominal 68\%. Where the previous work has demonstrated that foundation model embeddings can be used for spatial gap-filling, it was not validated for temporal gap-filling. If naive uncertainty estimation methods are unreliable in the spatial case, they are less applicable for temporal interpolation where the ground truth may have changed between training observations. Conformal prediction has recently been proposed as a distribution-free alternative for uncertainty quantification in Earth observation \citep{Singh2024ConformalPredictionEO, Valle2023ConformalLULC}, but provides only marginal coverage guarantees over the full test distribution. For monitoring applications where reliability in specific subpopulations such as actively disturbed forests is the primary concern, conditional calibration is required. 

The practical utility of temporal gap-filling is illustrated by GEDI's recent operational history, and more broadly for any data sparse in time-space dimensions, which is common in ecological sampling. The instrument entered hibernation on the ISS from March 2023 through April 2024, creating a 13-month gap in the global LIDAR record. Deforestation, fire, and degradation events that occurred during this period were not directly observed. Reconstructing biomass for such gap periods requires methods that interpolate reliably across time with trustworthy uncertainty bounds, particularly in regions experiencing active disturbance, where the assumption of temporal continuity is weakest.

Here, we test whether the Attentive Neural Process (ANP) framework previously introduced for spatial biomass interpolation by \citet{young2026interpolation} can be extended to jointly sparse spatiotemporal settings, where observations are irregular in both space and time and the underlying biomass field may be non-stationary due to disturbance. Temporal coordinates are concatenated alongside spatial ones as input dimensions over which the neural process interpolates, treating space and time symmetrically. If this unified treatment produces calibrated predictions, it empirically validates a form of space-for-time substitution, where observations from other years at nearby locations are informative about biomass at held-out times, just as observations from nearby locations at other times support spatial gap-filling. We evaluate this across three regions with contrasting ecological conditions and disturbance regimes, stratifying results by disturbance intensity to test whether calibration holds where the stationarity assumption is weakest.

Our three study regions are selected to span a range of biomass levels, disturbance processes, and ecological settings:
\begin{itemize}
    \item \textbf{Guaviare, Colombia} (2$^\circ$--3$^\circ$N, 72$^\circ$--73$^\circ$W): a tropical lowland deforestation frontier where expansion drives progressive forest clearing, creating a mosaic of intact forest corridors and agricultural land.
    \item \textbf{Ucayali, Peru} (10$^\circ$--11$^\circ$S, 74$^\circ$--75$^\circ$W): dense Amazonian rainforest subject to agricultural conversion and selective logging, with high baseline biomass.
    \item \textbf{Queensland, Australia} (26$^\circ$--27$^\circ$S, 144$^\circ$--145$^\circ$E): arid and semi-arid woodland with low standing biomass, providing a contrasting low-signal environment to assess whether the approach generalizes beyond tropical forests.
\end{itemize}

\section{Methods}

\subsection{Data sources and preprocessing}

We use GEDI Level 4A (L4A) AGBD estimates retrieved via the gediDB library \citep{besnard2025gediDB}, applying quality filtering on beam sensitivity, surface detection, and elevation agreement with TanDEM-X (see \citet{young2026interpolation} for filtering criteria and thresholds). The study period spans 2019--2023 inclusive. We treat GEDI L4A estimates as the target variable and apply log transformation to stabilize variance, following standard practice for biomass modeling \citep{Chave2014}. Values exceeding 500~Mg/ha ($<$1\% of observations) are excluded as likely instrumental artifacts \citep{sialelli2025agbd, Carreiras2017Coverage}.

For each GEDI footprint, we extract frozen embeddings from Tessera \citep{feng2025tessera}, a transformer-based remote sensing foundation model pretrained on Sentinel-1 (SAR) and Sentinel-2 (multispectral) imagery at global scale. Tessera belongs to a growing family of geospatial foundation models \citep{astruc2025anysat, danish2026terrafm, satmae2022, Szwarcman2026prithvi, brown2025alphaearth} that learn transferable representations from large-scale satellite data. We use Tessera because it jointly encodes Sentinel-1 and Sentinel-2 as temporal embeddings, aligning naturally with our spatiotemporal interpolation task. Tessera produces 128-dimensional embeddings at 10m resolution. We extract 3$\times$3 pixel patches (30m context) centered on each footprint, comparable to the GEDI footprint diameter (25m). Spatial coordinates are normalized to $[0,1]$ and biomass values are log-transformed and normalized to $[0,1]$ following \citet{young2026interpolation}. The model's task is therefore not to infer biomass from temporal context alone, but to transfer a learned embedding-to-biomass mapping to a year lacking LIDAR labels, with the model architecture providing calibrated uncertainty over this transfer.

\subsection{Attentive Neural Process architecture}

We use an Attentive Neural Process (ANP) \citep{kim2018attentive}, which extends the Conditional Neural Process \citep{garnelo2018conditional} with cross attention for context aggregation. Unlike standard supervised learning, which learns a single function mapping inputs to outputs, Neural Processes learn to produce a different predictive distribution for each set of context observations. Given a set of observed context points $C = \{(x_c, y_c)\}$ and target locations $\{x_t\}$, the model outputs a Gaussian predictive distribution $\mathcal{N}(\mu_t, \sigma_t^2)$ at each target, where the parameters depend on both the target's input features and the context set. This conditioning on context is what enables spatially and temporally adaptive uncertainty.

The architecture consists of five components. An \textit{embedding encoder} (3-layer CNN with residual connections) processes each 3$\times$3$\times$128 Tessera patch into a 1024-dimensional feature vector. A \textit{context encoder} (3-layer MLP with layer normalization) maps each context observation comprising the feature vector, normalized coordinates, and observed AGBD into a representation vector. A \textit{deterministic path} uses multihead cross-attention (16 heads) to aggregate context information to each target location, weighting context points by their relevance based on spatial proximity and feature similarity. This acts as a learned interpolation kernel that adapts to local data density and landscape structure. A \textit{stochastic latent path} summarizes the context set into a global latent distribution $\mathcal{N}(\mu_z, \sigma_z)$ via mean pooling, capturing function-level uncertainty not explained by the deterministic path. At inference, samples from this distribution (via the reparameterization trick \citep{kingma2014vae}) modulate predictions, increasing variance when context points are sparse or contradictory. Finally, a \textit{decoder} MLP combines the deterministic and stochastic representations to output the parameters $(\mu_t, \sigma_t^2)$ of the predictive Gaussian at each target. Complete architectural specifications are provided in \citet{young2026interpolation}.

The model is trained by the evidence lower bound (ELBO) objective:
\begin{equation}
    \mathcal{L} = -\mathbb{E}_{q(z|C,T)}[\log p(y_t | x_t, z, C)] + \beta \cdot \text{KL}[q(z|C, T) \| p(z|C)]
\end{equation}
where the first term is the negative log-likelihood of target observations under the predicted Gaussian, encouraging accurate predictions with appropriately scaled uncertainties, and the second term regularizes the latent space. We use $\beta$-annealing \citep{higgins2017betavae} to prevent posterior collapse.

Training is episodic. Each iteration samples a geographic tile, randomly partitions its GEDI observations into disjoint context and target sets (context ratio sampled uniformly from $[0.3, 0.7]$), and optimizes the ELBO by comparing predicted target distributions to held-out observations. This procedure is key for calibration because the model always predicts at locations it has not observed during each episode, it learns that uncertainty should reflect the density and consistency of nearby context points. Variable context set sizes further encourage robust uncertainty estimation across different observation densities. This episodic meta-learning structure is what enables the model to produce calibrated prediction intervals, in contrast to standard methods that learn a fixed input-output mapping without conditioning on local observations.

\subsection{Spatiotemporal extension}

We extend the ANP with explicit temporal awareness. Each observation is associated with a spatial coordinate $\mathbf{x}_{\text{loc}} = [\text{lon}, \text{lat}]$ and a temporal encoding:
\begin{equation}
    \mathbf{x}_{\text{time}} = \left[ \sin\left(\frac{2\pi d}{365}\right),\; \cos\left(\frac{2\pi d}{365}\right),\; \tau \right]
\end{equation}
where $d$ is the day of year (capturing seasonality) and $\tau \in [0,1]$ is the normalized timestamp relative to the full study period (capturing inter-annual position). The concatenated spatiotemporal coordinate $[\mathbf{x}_{\text{loc}}, \mathbf{x}_{\text{time}}]$ replaces the purely spatial coordinate. All other architectural components remain unchanged. This minimal modification treats space and time symmetrically as dimensions over which the neural process interpolates.

\subsection{Temporal holdout design}

We designate 2021 as the held-out test year. The model is trained exclusively on GEDI observations from $\{2019, 2020, 2022, 2023\}$. During evaluation, context observations are drawn only from training years and the model must predict 2021 biomass without any same-year observations. This simulates reconstructing biomass for a period that falls between GEDI operation periods.

\subsection{Spatiotemporal cross-validation}

The study region is partitioned into 0.1$^\circ \times$ 0.1$^\circ$ geographic tiles. For each experimental seed, tiles are randomly assigned to train (70\%), validation (15\%), and test (15\%) sets. A spatial buffer (0.1$^\circ$, approximately 11km) excludes all training and validation tiles adjacent to test tiles, preventing information leakage via spatial autocorrelation \citep{Roberts2017, ploton2020spatial, RejouMechain2014}. Predictions in test tiles are therefore from locations that are both spatially and temporally separated from training data. We use 10 random seeds per experiment.

\subsection{Disturbance stratification}

To evaluate performance specifically in areas undergoing forest change, we stratify the test set by disturbance intensity at tile level. The expected biomass $\bar{y}_{\text{exp}}$ is the mean AGBD across pre-event (2019--2020) and post-event (2022--2023) years. Disturbance intensity is the relative deviation of test-year biomass:
\begin{equation}
    \delta = \frac{\bar{y}_{\text{exp}} - \bar{y}_{2021}}{\bar{y}_{\text{exp}}}
\end{equation}
Tiles are classified as \textit{Stable} ($\delta < 0.1$), \textit{Moderate} ($0.1 \leq \delta \leq 0.3$), or \textit{Disturbed} ($\delta > 0.3$). At the shot densities in our study regions (typically $>$1000 shots per tile), tile-level means are robust to sampling variability, ensuring that high $\delta$ values reflect genuine biomass change.

Because disturbance events are sparse and unevenly distributed across random seeds, we compute pooled stratified metrics. Predictions and targets from all seeds are concatenated, and $R^2$ is calculated over this pooled set for each stratum:
\begin{equation}
    R^2_{\text{pooled}}(S) = 1 - \frac{\sum_{(y, \hat{y}) \in \mathcal{D}_{\text{pool}} \cap S} (y - \hat{y})^2}{\sum_{(y, \hat{y}) \in \mathcal{D}_{\text{pool}} \cap S} (y - \bar{y}_S)^2}
\end{equation}

\subsection{Baselines}

We compare the ANP against two baselines representing current practice for uncertainty-aware biomass estimation:

\textbf{Quantile Random Forest} (QRF) \citep{breiman2001random}: Rather than returning the mean prediction across trees, QRF retains the full distribution of training samples that reach each leaf node, estimating conditional quantiles from these empirical distributions. This provides prediction intervals that adapt to heteroscedastic structure in the data.

\textbf{XGBoost with quantile regression} (XGB) \citep{chen2016xgboost}: We train separate models to estimate the 16th and 84th conditional percentiles using pinball loss \citep{koenker1978regression}, approximating $\pm 1\sigma$ intervals. Quantile regression directly targets prediction intervals rather than deriving them from ensemble variance.

Both baselines use the same input features (concatenated normalized coordinates and embedding patches) and cross-validation procedure as the ANP.

\subsection{Evaluation}

We evaluate predictive accuracy via log-space $R^2$ (primary metric), log-space RMSE, and linear-space RMSE and MAE after back-transformation to Mg/ha units. Uncertainty calibration is assessed via standardized residuals $z = (y_{\text{true}} - y_{\text{pred}}) / \sigma_{\text{pred}}$, which should follow $\mathcal{N}(0,1)$ if uncertainties are well-calibrated. We report the $Z$-score mean (ideal: 0.0, indicating unbiased predictions) and $Z$-score standard deviation (ideal: 1.0; values $> 1$ indicate overconfident intervals, $< 1$ underconfident). We additionally report prediction interval coverage at $1\sigma$ (nominal 68\%) and $2\sigma$ (nominal 95\%). Together, these metrics distinguish models with accurate point predictions but unreliable uncertainty from models with trustworthy prediction intervals.

\section{Results}

\subsection{Global performance}

Table~\ref{tab:global} presents global accuracy and calibration metrics across all three study regions. The ANP achieves the highest log-space $R^2$ in all regions (0.75 in Guaviare, 0.50 in Queensland, 0.42 in Ucayali), with gains most pronounced in high biomass tropical sites. Linear-space errors are comparable across methods, with XGBoost achieving marginally lower RMSE in Queensland and Ucayali while the ANP leads in log-space metrics that better reflect relative prediction quality across the full biomass range.

\begin{table}[htbp]
\centering
\caption{Performance of spatiotemporal ANP versus baselines across three study regions (mean $\pm$ std over 10 seeds). Train years: 2019--2023 excluding 2021; test year: 2021.}
\label{tab:global}
\begin{tabular}{lccc}
\toprule
\textbf{Metric} & \textbf{QRF} & \textbf{XGB} & \textbf{ANP} \\
\midrule
\multicolumn{4}{l}{\textbf{Guaviare, Colombia}} \\
\midrule
\multicolumn{4}{l}{\textit{Accuracy}} \\
Log $R^2$ & $0.66 \pm 0.05$ & $0.70 \pm 0.05$ & $\mathbf{0.75 \pm 0.04}$ \\
Log RMSE & $0.230 \pm 0.018$ & $0.213 \pm 0.021$ & $\mathbf{0.196 \pm 0.019}$ \\
Linear RMSE (Mg/ha) & $51.1 \pm 4.5$ & $48.3 \pm 4.9$ & $\mathbf{48.3 \pm 5.4}$ \\
Linear MAE (Mg/ha) & $31.8 \pm 4.1$ & $29.3 \pm 4.3$ & $\mathbf{28.1 \pm 4.4}$ \\
\addlinespace
\multicolumn{4}{l}{\textit{Uncertainty Calibration}} \\
$1\sigma$ Coverage (68\%) & $74.8 \pm 2.3$ & $70.6 \pm 2.9$ & $\mathbf{77.2 \pm 2.6}$ \\
$2\sigma$ Coverage (95\%) & $89.8 \pm 1.2$ & $88.1 \pm 1.7$ & $\mathbf{92.9 \pm 1.6}$ \\
$Z$-Score Mean (0.0) & $-0.17 \pm 0.10$ & $-0.26 \pm 0.12$ & $\mathbf{-0.06 \pm 0.10}$ \\
$Z$-Score Std (1.0) & $1.32 \pm 0.11$ & $5.67 \pm 4.67$ & $\mathbf{1.19 \pm 0.17}$ \\
\midrule
\multicolumn{4}{l}{\textbf{Queensland, Australia}} \\
\midrule
\multicolumn{4}{l}{\textit{Accuracy}} \\
Log $R^2$ & $0.43 \pm 0.03$ & $0.49 \pm 0.04$ & $\mathbf{0.50 \pm 0.06}$ \\
Log RMSE & $0.123 \pm 0.004$ & $0.116 \pm 0.004$ & $\mathbf{0.115 \pm 0.006}$ \\
Linear RMSE (Mg/ha) & $8.3 \pm 3.1$ & $\mathbf{7.8 \pm 3.2}$ & $\mathbf{7.8 \pm 2.9}$ \\
Linear MAE (Mg/ha) & $3.8 \pm 0.3$ & $\mathbf{3.6 \pm 0.3}$ & $\mathbf{3.6 \pm 0.3}$ \\
\addlinespace
\multicolumn{4}{l}{\textit{Uncertainty Calibration}} \\
$1\sigma$ Coverage (68\%) & $68.5 \pm 1.5$ & $63.8 \pm 2.5$ & $\mathbf{70.5 \pm 3.9}$ \\
$2\sigma$ Coverage (95\%) & $92.8 \pm 1.0$ & $92.9 \pm 1.4$ & $\mathbf{93.9 \pm 1.9}$ \\
$Z$-Score Mean (0.0) & $\mathbf{-0.15 \pm 0.08}$ & $-0.16 \pm 0.09$ & $-0.20 \pm 0.13$ \\
$Z$-Score Std (1.0) & $1.08 \pm 0.04$ & $1.32 \pm 0.33$ & $\mathbf{1.01 \pm 0.09}$ \\
\midrule
\multicolumn{4}{l}{\textbf{Ucayali, Peru}} \\
\midrule
\multicolumn{4}{l}{\textit{Accuracy}} \\
Log $R^2$ & $0.32 \pm 0.09$ & $0.39 \pm 0.07$ & $\mathbf{0.42 \pm 0.14}$ \\
Log RMSE & $0.169 \pm 0.012$ & $0.160 \pm 0.011$ & $\mathbf{0.157 \pm 0.011}$ \\
Linear RMSE (Mg/ha) & $104.9 \pm 3.3$ & $\mathbf{99.5 \pm 2.6}$ & $104.0 \pm 3.7$ \\
Linear MAE (Mg/ha) & $83.4 \pm 3.4$ & $\mathbf{78.2 \pm 2.4}$ & $80.2 \pm 3.3$ \\
\addlinespace
\multicolumn{4}{l}{\textit{Uncertainty Calibration}} \\
$1\sigma$ Coverage (68\%) & $74.9 \pm 4.0$ & $67.2 \pm 2.3$ & $\mathbf{80.0 \pm 7.4}$ \\
$2\sigma$ Coverage (95\%) & $92.7 \pm 1.9$ & $90.6 \pm 1.5$ & $\mathbf{95.6 \pm 2.0}$ \\
$Z$-Score Mean (0.0) & $-0.30 \pm 0.13$ & $-0.32 \pm 0.11$ & $\mathbf{-0.09 \pm 0.09}$ \\
$Z$-Score Std (1.0) & $1.15 \pm 0.11$ & $1.34 \pm 0.08$ & $\mathbf{0.92 \pm 0.32}$ \\
\bottomrule
\end{tabular}
\end{table}

The ANP achieves $Z$-score standard deviation close to the ideal value of 1.0 across all regions (0.92--1.19), indicating that predicted uncertainties accurately reflect the empirical distribution of errors. XGBoost shows overconfident intervals, with $Z$-score standard deviation reaching 5.67 in Guaviare. QRF performs better on calibration than XGBoost but remains less well-calibrated than the ANP across all regions. These calibration patterns are consistent with those reported for the purely spatial case \citep{young2026interpolation}, indicating that the extension to temporal interpolation does not degrade uncertainty quality.

\begin{figure}[htbp]
    \centering
    \includegraphics[width=\textwidth]{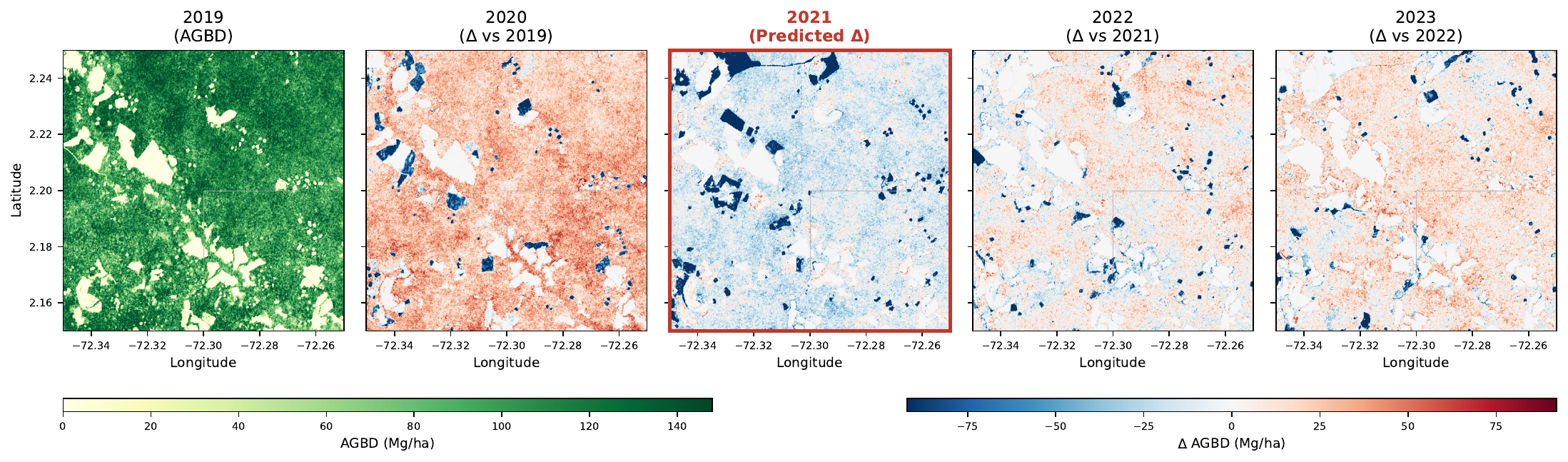}
    \caption{Temporal progression of biomass change for a tile in Guaviare, Colombia. Left: 2019 AGBD. Remaining panels show year-to-year change ($\Delta$ AGBD), with red indicating biomass gain and blue indicating loss. The central 2021 panel (red border) is the model's prediction from surrounding years.}
    \label{fig:progression}
\end{figure}

Figure~\ref{fig:progression} illustrates the spatiotemporal gap-filling for a representative tile in Guaviare. The leftmost panel shows the 2019 baseline AGBD, with high biomass (green) in forest patches and low biomass in cleared agricultural land. Subsequent panels show year-to-year biomass change, with the central 2021 panel (red border) fully interpolated using the model. The predicted 2021 change map shows biomass loss concentrated along forest-agriculture boundaries.

\subsection{Disturbance-stratified performance}

Table~\ref{tab:stratified} presents the stratified analysis. Performance is broken down by disturbance intensity across all three regions.

\begin{table}[htbp]
\centering
\caption{Disturbance-stratified performance (pooled across 10 seeds). Disturbance strata defined by relative biomass deviation $\delta$ between the held-out test year and surrounding training years.}
\label{tab:stratified}
\resizebox{\textwidth}{!}{%
\begin{tabular}{lccc|ccc|ccc}
\toprule
& \multicolumn{3}{c|}{\textbf{Guaviare}} & \multicolumn{3}{c|}{\textbf{Queensland}} & \multicolumn{3}{c}{\textbf{Ucayali}} \\
\textbf{Metric} & \textbf{QRF} & \textbf{XGB} & \textbf{ANP} & \textbf{QRF} & \textbf{XGB} & \textbf{ANP} & \textbf{QRF} & \textbf{XGB} & \textbf{ANP} \\
\midrule
\multicolumn{10}{l}{\textit{Stable ($\delta < 0.1$)}} \\
Log $R^2$ & $0.642$ & $0.690$ & $\mathbf{0.731}$ & $0.456$ & $0.510$ & $\mathbf{0.535}$ & $0.383$ & $0.439$ & $\mathbf{0.463}$ \\
Log RMSE & $0.238$ & $0.222$ & $\mathbf{0.207}$ & $0.121$ & $0.115$ & $\mathbf{0.112}$ & $0.160$ & $0.152$ & $\mathbf{0.149}$ \\
$Z$-Score Mean & $-0.16$ & $-0.26$ & $\mathbf{-0.04}$ & $\mathbf{-0.08}$ & $-0.09$ & $-0.12$ & $-0.24$ & $-0.25$ & $\mathbf{-0.04}$ \\
$Z$-Score Std & $1.37$ & $5.34$ & $\mathbf{1.28}$ & $1.09$ & $1.39$ & $\mathbf{1.02}$ & $1.11$ & $1.30$ & $\mathbf{0.91}$ \\
\midrule
\multicolumn{10}{l}{\textit{Moderate ($0.1 \leq \delta \leq 0.3$)}} \\
Log $R^2$ & $0.721$ & $0.770$ & $\mathbf{0.810}$ & $0.416$ & $0.471$ & $\mathbf{0.476}$ & $0.138$ & $0.223$ & $\mathbf{0.238}$ \\
Log RMSE & $0.206$ & $0.187$ & $\mathbf{0.170}$ & $0.126$ & $0.120$ & $\mathbf{0.119}$ & $0.198$ & $0.188$ & $\mathbf{0.187}$ \\
$Z$-Score Mean & $-0.22$ & $-0.27$ & $\mathbf{-0.09}$ & $\mathbf{-0.23}$ & $-0.25$ & $-0.30$ & $-0.53$ & $-0.58$ & $\mathbf{-0.31}$ \\
$Z$-Score Std & $1.20$ & $10.12$ & $\mathbf{1.03}$ & $1.08$ & $1.41$ & $\mathbf{1.02}$ & $1.32$ & $1.50$ & $\mathbf{1.13}$ \\
\midrule
\multicolumn{10}{l}{\textit{Disturbed ($\delta > 0.3$)}} \\
Log $R^2$ & $0.635$ & $0.656$ & $\mathbf{0.748}$ & $0.296$ & $\mathbf{0.390}$ & $0.375$ & $-0.165$ & $0.087$ & $\mathbf{0.201}$ \\
Log RMSE & $0.207$ & $0.201$ & $\mathbf{0.172}$ & $0.114$ & $\mathbf{0.106}$ & $0.107$ & $0.246$ & $0.217$ & $\mathbf{0.203}$ \\
$Z$-Score Mean & $\mathbf{-0.13}$ & $-0.38$ & $-0.19$ & $-0.30$ & $\mathbf{-0.29}$ & $-0.38$ & $-1.00$ & $-0.94$ & $\mathbf{-0.46}$ \\
$Z$-Score Std & $1.07$ & $13.03$ & $\mathbf{0.94}$ & $\mathbf{0.97}$ & $1.11$ & $0.92$ & $1.40$ & $1.70$ & $\mathbf{1.32}$ \\
\bottomrule
\end{tabular}%
}
\end{table}

Across all regions, the ANP's advantage over baselines grows with disturbance intensity. In Guaviare, the gap in log $R^2$ between the ANP and XGBoost widens from 0.04 in stable tiles to 0.09 in disturbed tiles. In Ucayali, the QRF achieves negative $R^2$ ($-0.17$) in the disturbed stratum, meaning its predictions are worse than the stratum mean; XGBoost reaches 0.09; while the ANP maintains $R^2 = 0.20$. Though 0.20 represents limited explanatory power, it indicates the model captures some of the biomass signal even in areas of severe forest loss which is a regime where baseline methods fail.

Queensland presents a partial exception as XGBoost achieves the highest $R^2$ in the disturbed stratum (0.39 vs.\ 0.38 for the ANP). However, XGBoost's calibration is overconfident in this stratum ($Z$-score std = 1.11 vs.\ 0.92 for the ANP), meaning its point predictions are better but its uncertainty intervals are less reliable.

Calibration remains the ANP's most consistent advantage across strata. In the disturbed tiles of Guaviare, XGBoost's $Z$-score standard deviation reaches 13.03. XGBoost's prediction intervals in this regime would obtain intervals that are an order of magnitude too narrow relative to actual errors. The ANP maintains $Z$-score standard deviation of 0.94 in the same stratum, close to the ideal 1.0.

\section{Discussion}

The disturbance-stratified analysis reveals that the ANP degrades more gracefully than baselines in the conditions that matter most for forest monitoring. In the disturbed stratum of Ucayali, which is a dense Amazonian forest undergoing active conversion, the gap between the ANP and standard methods is largest. This is consistent with the spatial context awareness of the model architecture as the model conditions predictions on nearby observations across time, allowing it to detect inconsistencies between pre- and post-disturbance context that signal change. Ensemble methods, which treat each prediction independently given input features, lack this mechanism. The lower predictive performance in Ucayali likely reflects, in part, sensor-level limitations. Sentinel-1 C-band backscatter and Sentinel-2 optical reflectance both saturate at biomass densities well below the levels typical of dense Amazonian forest, imposing an information ceiling on any model derived from these inputs. Future integration of longer-wavelength SAR sensors such as NISAR \citep{rosen2025nisar} or the BIOMASS \citep{letoan2011biomass} mission, which penetrate deeper into the canopy structure, could improve retrievals in high-biomass tropical settings.

Because the ANP conditions on a set of context observations and predicts at arbitrary query locations, the trained model defines a continuous predictive surface over space and time such that any (latitude, longitude, date) tuple within the training domain yields a full posterior predictive distribution over biomass. This property has direct methodological consequences. Temporal gap-filling becomes an instance of the same interpolation the model performs spatially, rather than a separate task requiring different machinery. The calibration of these predictions matters because downstream applications whether change detection, carbon accounting, or anomaly flagging depend on prediction intervals reflecting actual error distributions. A model with $Z$-score standard deviation of 13 in disturbed areas does not just have wide uncertainty, it has uncertainty estimates that render any interval-based inference unreliable. That the ANP maintains $Z$-score standard deviation near 1.0 across strata means its uncertainty can be used for downstream tasks without post-hoc recalibration.

The inclusion of Queensland as an arid, low-biomass environment with different vegetation dynamics demonstrates that the approach is not restricted to tropical forests. While absolute performance is lower in dense tropical sites (reflecting the inherent difficulty of high-biomass estimation), the calibration advantage under disturbance are consistent across biomes. This suggests the method may have potential as well for being applied to diverse monitoring contexts, from savanna degradation to dryland carbon accounting.

Our results suggest that the 2023-2024 GEDI hibernation gap is amenable to reconstruction via spatiotemporal neural process interpolation as the model maintains calibrated uncertainty when interpolating across a held-out year even under active disturbance, which is the condition that makes gap-filling most difficult and most valuable. The method should generalize beyond GEDI, since any spaceborne mission with intermittent coverage will produce temporal gaps, and the combination of foundation model embeddings with probabilistic spatiotemporal interpolation provides an approach to bridging them.

The feasibility of this approach depends on the availability of foundation model embeddings that capture time-varying land surface state. Tessera provides temporal embeddings by jointly encoding Sentinel-1 and Sentinel-2 imagery acquired at the time of prediction, meaning the embedding reflects the landscape as it exists at a given date rather than as a static summary. This temporal awareness is what allows the model to detect change between training and test periods through the input features themselves, independent of the GEDI observations. Many existing geospatial foundation models produce fixed per-location representations that do not vary with acquisition date, which would preclude the spatiotemporal interpolation demonstrated here. As the geospatial foundation model ecosystem matures, temporal encoding should be considered a key capability for downstream tasks that require inference across time, not only for biomass gap-filling but any application where the quantity of interest is non-stationary.

Space-for-time substitution is broadly distrusted in ecology and remote sensing because it assumes stationarity, and disturbance is the condition under which stationarity breaks down \citep{damgaard2019critique}. Our results demonstrate that spatiotemporal interpolation of GEDI biomass estimates is nonetheless feasible, and that the calibration advantage of ANPs over ensemble methods, established for the spatial case \citep{young2026interpolation}, extends to the temporal domain. The temporal holdout design tests a harder inference problem than spatial gap-filling as the model must reconstruct biomass for a year it has never observed, where the land surface may have changed between training periods. Despite this, the ANP maintains near-ideal uncertainty calibration across three diverse ecosystems not because the stationarity assumption holds but because the model reports when it does not. Calibration ensures that reported uncertainties are trustworthy for downstream decision-making; whether those uncertainties are sufficiently small for a given application depends on observation density and landscape complexity, not the choice of model. When calibrated uncertainty is too large for a given decision threshold, we see this as informative as it identifies where additional observations would be most valuable, which could drive informed sampling strategies such as active learning. A calibrated model that reports high uncertainty in a region of interest directs field campaigns or future LIDAR acquisitions toward locations where they will most reduce decision-relevant uncertainty, closing the loop between prediction and data collection. An overconfident model forecloses this feedback loop by masking the locations where additional data is most needed.

Several limitations warrant discussion. The disturbance stratification operates at tile level ($\sim$11\,km), which is coarser than individual disturbance events. Footprint-level disturbance detection would require denser temporal sampling than GEDI currently provides but could be ameliorated by the upcoming EDGE mission. The temporal encoding provides the model with information about seasonality and inter-annual position and for it to be possible at all to query the model at arbitrary space-time for inferred estimates, but the primary signal for detecting change between training and test periods comes from the embeddings, which capture land surface state at the time of prediction independently of GEDI observations. Isolating the marginal contribution of the temporal coordinates relative to the embeddings, or if there are better ways to represent temporality, would be a direction for future work, though it does not affect the practical finding that spatiotemporal gap-filling with calibrated uncertainty is achievable.

\section{Data Availability}

GEDI-L4A data is publicly available through \href{https://doi.org/10.3334/ORNLDAAC/2056}{NASA EarthData}, with gediDB \citep{besnard2025gediDB} used for access and organization. Tessera \citep{feng2025tessera} embeddings are publicly available through the Python package \href{https://github.com/ucam-eo/geotessera}{geotessera}.



\section{Author Contributions}

RY: Conceptualization; Data curation; Formal analysis; Investigation; Methodology; Validation; Visualization; Software; Writing -- original draft; Writing -- review and editing.

SK: Conceptualization; Supervision; Writing -- review and editing.
    
\section{Funding}

This work was supported by the Taiwan Cambridge Scholarship from the Cambridge Trust and by funding from Dr. Robert Sansom.

\bibliographystyle{elsarticle-num-names} 
\bibliography{cas-refs}

\appendix

\end{document}